\pdfoutput=1

\documentclass[11pt]{article}

\usepackage[final]{acl}

\usepackage{times}
\usepackage{latexsym}
\usepackage{amsmath}
\usepackage{amssymb}
\usepackage[T1]{fontenc}

\usepackage[utf8]{inputenc}

\usepackage{microtype}
\usepackage{adjustbox}
\usepackage{inconsolata}

\usepackage{graphicx}

\title{Under-resourced studies of under-resourced languages: lemmatization and POS-tagging with LLM annotators for historical Armenian, Georgian, Greek and Syriac}

\author{
  \textbf{Chahan Vidal-Gorène\textsuperscript{1,2}},
  \textbf{Bastien Kindt\textsuperscript{3}},
  \textbf{Florian Cafiero\textsuperscript{2}}
\\
\\
  \textsuperscript{1}LIPN, CNRS UMR 7030, France \\
  \textsuperscript{2}École nationale des chartes, Université Paris-Sciences-et-Lettres (PSL), France \\
  \textsuperscript{3}Université catholique de Louvain, Belgium
\\
  \small{
    \textbf{Correspondence:} \href{mailto:chahan.vidal-gorene@chartes.psl.eu}{chahan.vidal-gorene@chartes.psl.eu}
  }
}

\begin{document}
\maketitle
\begin{abstract}
Low-resource languages pose persistent challenges for Natural Language Processing tasks such as lemmatization and part-of-speech (POS) tagging. This paper investigates the capacity of recent large language models (LLMs), including GPT-4 variants and open-weight Mistral models, to address these tasks in few-shot and zero-shot settings for four historically and linguistically diverse under-resourced languages: Ancient Greek, Classical Armenian, Old Georgian, and Syriac. Using a novel benchmark comprising aligned training and out-of-domain test corpora, we evaluate the performance of foundation models across lemmatization and POS-tagging, and compare them with PIE, a task-specific RNN baseline. Our results demonstrate that LLMs, even without fine-tuning, achieve competitive or superior performance in POS-tagging and lemmatization across most languages in few-shot settings. Significant challenges persist for languages characterized by complex morphology and non-Latin scripts, but we demonstrate that LLMs are a credible and relevant option for initiating linguistic annotation tasks in the absence of data, serving as an effective aid for annotation.
\end{abstract}

\section{Introduction}

While not always surpassing state-of-the-art specialized methods, Large Language Models demonstrated remarkable zero-shot performance in tasks such as POS-tagging for languages such as Arabic \cite{alyafeai2023taqyim} and Armenian \cite{vidal2024cross}. In related tasks such as NER, recent advancements for low-resource languages have demonstrated promising results through knowledge transfer from large pre-trained language models (PLMs). In particular, a recent study \cite{tolegen2024enhancing} showcased significant improvements in Kazakh NER by leveraging multilingual models such as XLM-RoBERTa, achieving a 7\% increase in F1-score compared to previous approaches. However, their findings also revealed that few-shot learning scenarios remain a challenge, with multilingual PLMs struggling to achieve high performance when trained on limited data. Interestingly, their comparison with GPT-4 indicated that such large-scale generative models exhibit superior adaptability, even in low-data scenarios, suggesting the potential of these models to generalize across diverse linguistic settings.

This paper aims to build upon these insights by exploring how GPT-4 and similar foundation models can offer a more agile and effective approach to low-resource NLP, demonstrating their ability to bridge the gap between high- and low-resource languages with minimal fine-tuning effort.

\subsection{State of the Art}

Morphological annotation tasks such as lemmatization and POS tagging in under-resourced and historically diverse languages have been approached using rule-based systems, sequence-based neural architectures, and more recently, large language models.

For Armenian, \citet{vidal-gorene-etal-2020-recycling} demonstrate that RNN-based models trained on Eastern Armenian can be reused to annotate dialectal and Western varieties, outperforming rule-based baselines and showing high transferability. Similarly, in the context of Ancient Greek, \citet{kindt2022analyse} evaluate an RNN trained on patristic Greek to annotate historiographical prose, achieving over 97\% accuracy in both lemmatization and POS-tagging, with strong performance on ambiguous and unseen forms.

\citet{manjavacas2019improving} introduce a joint-learning architecture for lemmatization that combines character-level transduction with a sentence-level language modeling objective. Applied to non-standard historical languages such as Latin, Old French, the model surpasses baselines, especially on ambiguous forms, without relying on gold morphological annotations. It has then been applied to other historical languages like Old French and Classical French \cite{camps2021corpus, cafiero2019moliere, clerice2025wauchier} or 14th century Dutch \cite{creten2020linguistic}, and integrated into specialized annotation infrastructures \cite{clerice:hal-03606756}.

The feasibility of extending such models to typologically diverse languages is addressed by \citet{vidal2020lemmatization}, who apply a character-level RNN (PIE) to Classical Armenian, Old Georgian, and Syriac. Their models reach over 91\% accuracy in lemmatization and 92\% in POS tagging, using curated corpora derived from the GRE\textit{g}ORI project. Results confirm the adaptability of RNNs to languages with complex polylexical structures, although accuracy on ambiguous and unknown forms remains lower (e.g., 71.9\% F1 on unknown tokens in Classical Armenian). The authors emphasize the limitations of rule-based pipelines and the potential of neural models to handle diachronic and morphologically rich corpora written in the different languages of the Christian East.

More recently, \citet{vidal2024cross} benchmark RNNs, mDeBERTa, and GPT-4 across four Armenian varieties, including Classical and a rare spoken dialect. RNNs remain competitive for POS-tagging (F1 $>$ 0.98), while GPT-4 demonstrates strong generalization in zero- and few-shot setups, achieving F1 = 0.83 in lemmatization on unseen dialect data. Transformer models underperform on morphologically rich and low-resource dialects, particularly in low-data conditions.

Taken together, these studies highlight the importance of context-aware neural architectures for low-resource NLP. They also point to persistent challenges: heterogeneous annotation schemes, limited corpus coverage, and the need for robust cross-dialectal generalization.

\section{Datasets}

Our data sets comprise texts in Greek, Armenian, Georgian, and Syriac, representing the linguistic and cultural diversity of the Christian East. Each language represents a sub-dataset, divided into two parts: a 'Training corpus' (5,000 words) for model training, and an out-of-domain 'Test corpus' (300 words) for evaluation purposes (see Table \ref{tab:dataset_texts}).

Each text underwent preliminary linguistic analysis, including lemmatization and POS-tagging, tagged with a hybrid approach of rule-based and RNN models \cite{vidal2020lemmatization,kindt2022analyse}.

\subsection{Language families}

The languages in this dataset belong to different language families: Greek (Indo-European) with a rich inflectional system; Armenian (Indo-European) with a partially inflectional and agglutinative system; Georgian (Kartvelian) with a more distinctly agglutinative system; and Syriac (Semitic), which presents a non-concatenative, templatic morphology.
The texts were produced between the 4th and 15th centuries AD and cover a wide range of topics, including asceticism, epistolography, exegesis, hagiography, historiography, homiletics, patristics, pseudepigrapha, and theology. They are either original texts or ancient translations, all previously published in critical editions. This sample, though only partially representative, is highly varied and diversified.

\begin{table*}[ht]
\centering
\begin{tabular}{l|p{6.5cm}|p{6.5cm}}
\hline
\multicolumn{1}{c|}{} &
\multicolumn{1}{c|}{\textbf{Training Corpus}} &
\multicolumn{1}{c}{\textbf{Test Corpus}} \\
\hline
\textbf{Greek} &
  John Anagnostes, \textit{De Thessalonica Capta} (15th c. AD - historiography) &
  Gregory of Nazianzus, \textit{Homily 1. In Sanctum Pascha} (4th c. AD - patristics, homiletic) \\
\hline
\textbf{Armenian} &
  Evagrius, \textit{Letters} (13th c. AD - epistolography, ascetism) &
  Step'anos of Siwnik' (Dub.), \textit{The Genesis Commentary} (8th-9th c. AD - exegesis, theology) \\
\hline
\textbf{Georgian} &
  Anonymus, \textit{Conversion of the Kartli} (5th-9th c. AD - hagiography, historiography) &
  \textit{History of the Rechabites} by Zosimus (8th-10th c. AD - hagiography, historiography) \\
\hline
\textbf{Syriac} &
  Jacob of Serugh, \textit{Homilies} (5th-6th c. AD - patristics, homiletic) &
  \textit{History of the Rechabites} by Zosimus (Long Version) (4th c. AD - hagiography, pseudepigraphs) \\
\hline
\end{tabular}
\caption{Overview of training and test corpora: chronological and genre diversity.}
\label{tab:dataset_texts}
\end{table*}

\subsection{Annotation guidelines and tagsets}

\begin{figure}[ht!]
    \centering
    \includegraphics[width=\linewidth]{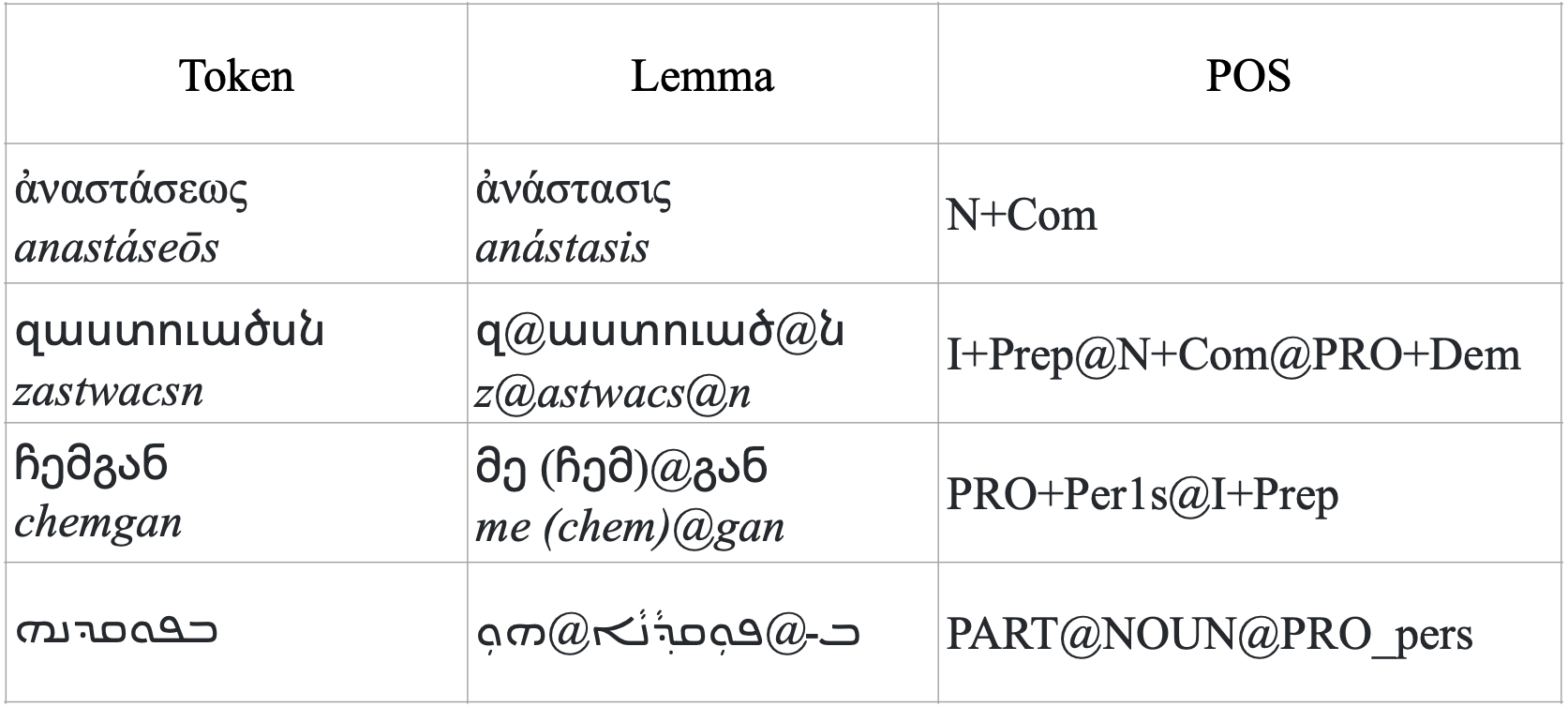}
    \caption{Annotation guidelines and tagset for Greek, Armenian, Georgian and Syriac, using \symbol{64} to split agglutinated and polylexical forms}
    \label{fig:lang_overview}
\end{figure}

The lemmatization and POS tagging follow the GRE\textit{g}ORI annotation guidelines \citet{gregori-online-corpus}, which use the @ symbol to discriminate the lexical elements constituting a polylexical form (crases of Greek; agglutinated forms of Georgian; prefixed and suffixed forms of Syriac, etc.). For instance, a word composed of multiple lemmas is annotated as lemma1@lemma2 with corresponding POS tags as pos1@pos2 (see Figure \ref{fig:lang_overview}).
Unlike standard annotation frameworks such as Universal Dependencies, this method provides a more granular representation by explicitly marking all constituent lemmas and their POS tags. However, it lacks direct equivalents in publicly available datasets and increases task complexity, making model training more challenging. Additionally, the GRE\textit{g}ORI tagset diverges from conventional tagsets in its categorization and structure, further complicating annotation and processing \cite{coulie2013lemmatization,automatique2022etiquettes,atas2022principles}.

The full tagset is provided in Table~\ref{tab:pos-list} in the appendix. These POS tags introduce additional complexity compared to standard tagsets, as some would typically be treated as morphological markers rather than POS categories (e.g., \texttt{PRO+Per1d}, \texttt{PRO+Per3s}, \texttt{PRO+Ref1s}). We retain the original annotations without normalization\footnote{For full annotation scheme guidelines, see Armenian \cite{automatique2022etiquettes}, Georgian \cite{coulie2013lemmatization}, Greek \cite{kindt:hal-01018202}, and Syriac \cite{atas2018concordance}.}.

\subsection{Polylexicality and Segmentation}
A critical challenge in our dataset, and particularly for Syriac, is the prevalence of polylexical forms, where a single graphical token corresponds to multiple syntactic units (e.g., a preposition attached to a noun). In the GRE\textit{g}ORI schema, these are annotated by splitting the form with an \texttt{@} symbol (e.g., \textit{lemma1@lemma2}).

As shown in Table~\ref{tab:polylexical_dist}, the density of these forms varies drastically across languages. Syriac exhibits an exceptionally high density, with nearly 42\% of tokens in the training set being polylexical, reflecting its Semitic morphology where prepositions and conjunctions are prefixed. In contrast, Greek and Georgian (in this specific annotation schema) show significantly lower rates of explicit segmentation.

\begin{table}[h]
\centering
\small
\setlength{\tabcolsep}{2.5pt} 
\resizebox{\columnwidth}{!}{%
\begin{tabular}{|l|rr|rr|rr|}
\hline
& \multicolumn{2}{c|}{\textbf{Train}} & \multicolumn{2}{c|}{\textbf{Test (In)}} & \multicolumn{2}{c|}{\textbf{Test (Out)}} \\
\textbf{Lang} & \textbf{Poly.} & \textbf{Simple} & \textbf{Poly.} & \textbf{Simple} & \textbf{Poly.} & \textbf{Simple} \\ \hline
\textbf{GRC} & 13 & 4705 & 2 & 298 & 1 & 301 \\
\textbf{HYE} & 669 & 4043 & 56 & 244 & 55 & 259 \\
\textbf{KAT} & 156 & 4551 & 5 & 295 & 5 & 300 \\
\textbf{SYC} & 1696 & 2296 & 144 & 156 & 140 & 161 \\ \hline
\end{tabular}%
}
\caption{Distribution of polylexical vs. simple forms across datasets.}
\label{tab:polylexical_dist}
\end{table}

\section{Tasks}

\subsection{Task definition: lemmatization and POS tagging}

We consider two classic sequence-labeling tasks: lemmatization and part-of-speech (POS) tagging. In each case, the problem can be formulated as follows. Given an input sequence of $n$ tokens $\mathbf{x} = \{x_1, x_2, \ldots, x_n\}$, we wish to predict a sequence of labels $\hat{\mathbf{y}} = \{\hat{y}_1, \hat{y}_2, \ldots, \hat{y}_n\}$ from an appropriate label set. For \emph{lemmatization}, each label $y_i$ is the canonical lemma form of token $x_i$. In \emph{POS tagging}, $y_i$ is assigned one of a finite set of syntactic categories (e.g., noun, verb).
A collection of input-label pairs forms our data set $\mathcal{D}$. We train a model with parameters $\boldsymbol{\theta}$ to maximize the conditional log-likelihood of the correct label sequences:
\[
\max_{\boldsymbol{\theta}} \sum_{(\mathbf{x}, \mathbf{y}) \in \mathcal{D}} \log P(\mathbf{y} \,\big\lvert\, \mathbf{x}; \boldsymbol{\theta}).
\]

The \textit{POS-tagging} label set includes a small number of NER-related categories (e.g., \texttt{N+Ant}, \texttt{N+Epi}, \texttt{N+Pat}, \texttt{N+Prop}, \texttt{N+Top}, \texttt{NAME}, \texttt{NAME\_ant}, \texttt{NAME\_top}, see Table~\ref{tab:pos-list} in the appendix), but these are too infrequent in the data to support separate evaluation.

\subsection{Prompt Engineering}\label{sec:prompts}

To adapt general-purpose LLMs to the specificities of the GRE\textit{g}ORI annotation schema, we employed a structured prompting strategy based on the COSTAR framework \citep{teo2023prompt}. This approach decomposes the prompt into six components: Context, Objective, Style, Tone, Audience, and Response.

Given the non-standard nature of our target tagset (which includes morphological markers fused with POS tags), standard instruction tuning was insufficient. We therefore implemented two critical constraints within the prompt design (see Figure~\ref{fig:prompt_structure}):

\begin{enumerate}
    \item \textbf{Tagset Injection:} We explicitly injected the full list of valid POS tags ($|\mathcal{T}| \approx 60$) into the prompt context. This acts as a strict in-context constraint, drastically reducing the likelihood of the model hallucinating invalid tags or reverting to Universal Dependencies tags.
    \item \textbf{Segmentation Guidance:} To address the challenge of polylexicality (e.g., Syriac prefixed prepositions), the prompt includes a specific instruction and a concrete example demonstrating the use of the \texttt{@} delimiter to split lemmata and tags (e.g., mapping the Greek \textit{tauton} to \textit{ho@autos}).
\end{enumerate}

For the \textbf{Few-Shot} settings ($k=5, 50, 500$), we appended $k$ aligned examples (token, lemma, POS) from the training set. These examples were selected sequentially from the start of the training corpus to maintain narrative coherence. Unlike random sampling, this sequential presentation allows the model to leverage sentence-level context for morphosyntactic disambiguation.

\begin{figure}[t!]
\centering
\small
\begin{adjustbox}{width=\linewidth}
\begin{tabular}{|p{0.95\linewidth}|}
\hline
\textbf{Context:} You are an expert in computational linguistics with a specialization in [LANGUAGE]... \\
\textbf{Objective:} Examine the provided content and assign lemma and POS tags... \\
\textbf{Format:} A .tsv table with three columns... \\
\textbf{Constraints:} \\
1. All words must be annotated. \\
2. Only use tags from the provided list: \\
\texttt{[INSERT FULL TAGSET LIST HERE]} \\
\textbf{Segmentation Rule:} \\
A form can be a combination of multiple tokens. Identify combinations with `@`. \\
\textit{Example:} [Language Specific Example of Split] \\
\hline
\textbf{Few-Shot Examples:} \\
form \hspace{1cm} lemma \hspace{1cm} pos \\
$w_1$ \hspace{1.1cm} $l_1$ \hspace{1.4cm} $p_1$ \\
... \\
$w_k$ \hspace{1.1cm} $l_k$ \hspace{1.4cm} $p_k$ \\
\hline
\textbf{Input text to process:} \\
\texttt{[TEST INPUT]} \\
\hline
\end{tabular}
\end{adjustbox}
\caption{Schematic representation of the prompt structure used for all languages.}
\label{fig:prompt_structure}
\end{figure}

\subsection{Decoding Strategy}

We utilized model-specific decoding parameters to optimize for stability and adherence to the tagging schema.

\begin{itemize}
    \item \textbf{Greedy Decoding ($\tau = 0.0$):} For \textit{GPT-4o}, we employed strictly deterministic decoding. These models demonstrated high adherence to constraints without requiring stochastic noise to avoid degeneration.
    
    \item \textbf{Low-Temperature Sampling ($\tau = 0.2$):} For the \textit{GPT-4o-mini}, \textit{Mistral Nemo}, and \textit{Mistral Large} experiments, we applied a slight temperature increase. We standardized the batch processing for these groups with $\tau = 0.2$. This setting served as a preventative measure against repetitive loops observed in preliminary experiments, ensuring robust generation across extensive test sets.
    
    \item \textbf{Reasoning Models:} For \textit{o1-mini}, we utilized the model's default decoding parameters to preserve the integrity of the internal chain-of-thought process.
\end{itemize}

\subsection{Baseline definition}

While Transformer-based encoders (e.g., mDeBERTa) generally define the state-of-the-art for high-resource languages, recent work on historical and dialectal varieties suggests they are not always the optimal baseline. \citet{vidal2024cross} demonstrated that for character-level tasks in Classical and Modern Armenian, the RNN-based PIE architecture consistently outperformed mDeBERTa, particularly in low-data regimes where Transformers struggle to generalize without extensive fine-tuning. Furthermore, from a pragmatic perspective, lightweight RNN architectures remain significantly more accessible to the digital humanities community, especially in the context of low-resource studies. RNN offer lower computational overhead and greater ease of deployment for non-technical experts compared to large Transformer pipelines. 

\begin{table}[ht]
\centering
\resizebox{\columnwidth}{!}{%
\begin{tabular}{ll}
\hline
\textbf{Component / Parameter} & \textbf{Configuration} \\
\hline
Encoder Architecture & 2-layer Bi-directional GRU \\
Hidden Size & 300 \\
Char Embeddings & 300 (2-layer RNN) \\
Word Embeddings & 0 (Disabled) \\
Lemma Decoder & Attentional (Character-level) \\
POS Decoder & CRF (Conditional Random Field) \\
Dropout / Batch Size & 0.25 / 50 \\
Optimization / LR & Adam (0.001) \\
LR Scheduler & Factor: 0.75 (Patience: 2) \\
Early Stopping & Patience: 3 epochs (Max 100) \\
\hline
\end{tabular}%
}
\caption{Hyperparameter configuration for the PIE supervised baseline.}
\label{tab:pie_params}
\end{table}

\begin{figure*}[ht!]
    \centering
    \includegraphics[width=\linewidth]{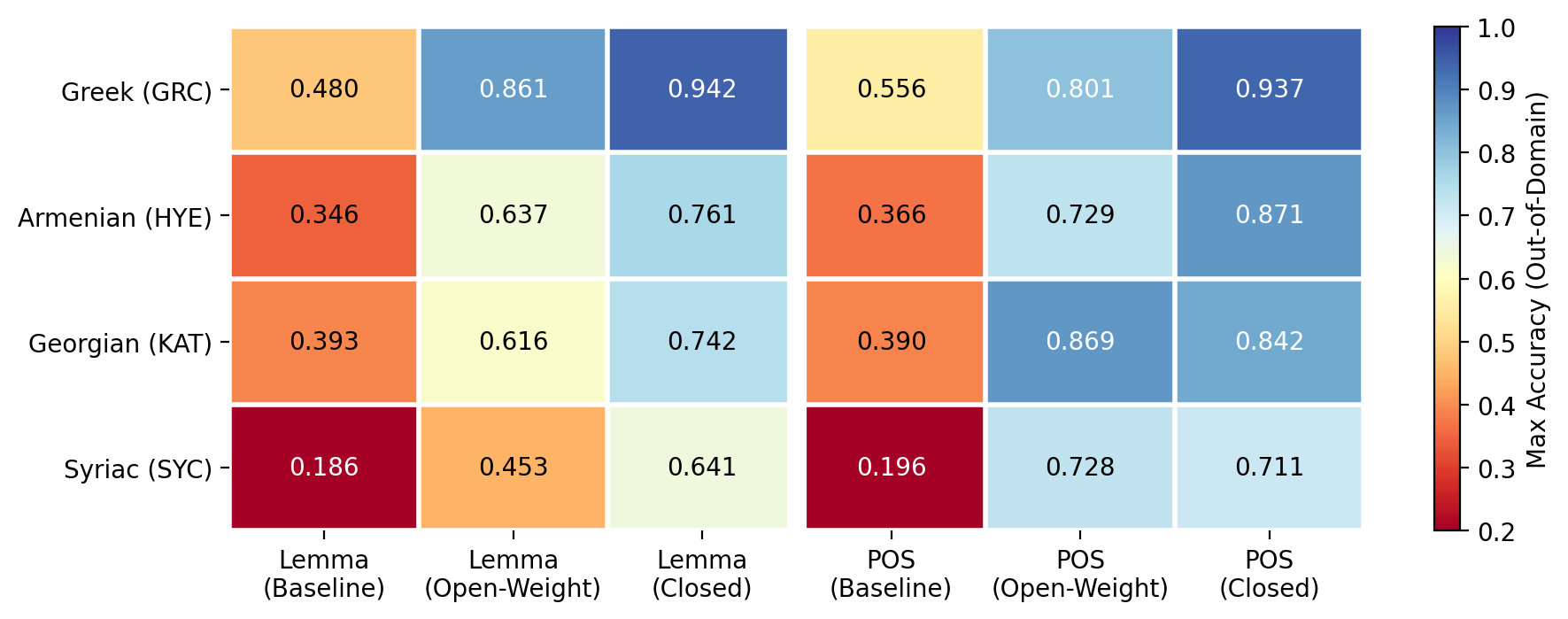}
    \caption{Out-of-domain accuracy for lemmatization and POS tagging across our four historical languages: comparing the supervised PIE baseline with best open-weight and closed LLM annotators; values report maximum accuracy by task and language.}
    \label{fig:heatmap}
\end{figure*}

Consequently, we retain PIE as our primary supervised baseline. PIE is a multi-task character-level RNN architecture specifically designed for the joint annotation of morphologically rich languages. For all languages in this study, we utilize a standardized configuration, summarized in Table \ref{tab:pie_params}, to ensure comparability and force the model to rely exclusively on character-level morphological patterns.

\section{Results}

Tables~\ref{tab:lemmatization-results} and~\ref{tab:pos-results} summarize our complete results, evaluated in both in-domain and out-of-domain scenarios. A simplified comparison is provided in figure ~\ref{fig:heatmap}.

\subsection{Lemmatization}

Overall, large language models, particularly GPT-4o, but even open-weight models like mistral-large, significantly outperform the traditional RNN-based model PIE across most scenarios, especially in low-shot conditions. For Greek, GPT-4o attains the highest accuracy with \textbf{0.933} (in-domain) and \textbf{0.942} (out-of-domain) at 500 shots, though mistral-large also demonstrates robust zero-shot performance (0.863 in-domain, 0.861 out-of-domain). For Armenian, GPT-4o achieves notable accuracy (\textbf{0.796} in-domain, \textbf{0.761} out-of-domain at 500 shots), while mistral-large remains competitive in medium-shot scenarios. Georgian results indicate consistent difficulties across models, although GPT-4o still maintains leading performance at \textbf{0.760} (in-domain, 500 shots). Syriac poses the greatest challenge overall; GPT-4o reaches a peak accuracy of \textbf{0.656} (in-domain), and open-weight models generally demonstrate moderate performances.

The PIE baseline performs substantially lower, illustrating limitations of traditional sequence labeling architectures when data availability is limited.

\begin{table*}[ht]
\centering
\resizebox{\textwidth}{!}{%
\begin{tabular}{|l|rr|l|rr|}
\hline
\textbf{Lang} & \multicolumn{2}{c|}{\textbf{Lemma Overlap (\%)}} & \textbf{Top 3 POS (Test Set)} & \multicolumn{2}{c|}{\textbf{Polylexical (@) \%}} \\
& Type & Token & & Train & Test \\
\hline
\textbf{GRC} & 47.6 & 62.9 & V (17.9\%), N+Com (17.5\%), I+Part (16.2\%) & 0.3 & 0.3 \\
\textbf{HYE} & 52.7 & 67.2 & V (18.8\%), I+Conj (17.8\%), N+Com (15.6\%) & 14.2 & 17.5 \\
\textbf{SYC} & 43.4 & 42.2 & NOUN (15.6\%), PART@NOUN (14.6\%), ADJ (11.3\%) & 42.5 & 46.5 \\
\textbf{KAT} & 62.9 & 76.4 & N+Com (22.3\%), V+Mas (20.3\%), I+Conj (17.4\%) & 3.3 & 1.6 \\
\hline
\end{tabular}}
\caption{Lexical overlap, dominant POS distributions, and density of polylexical forms across languages.}
\label{tab:lexical_analysis}
\end{table*}

\subsection{POS Tagging}

In POS-tagging, GPT-4o and mistral-large both show strong results, particularly in Greek, where GPT-4o reaches \textbf{0.963} (in-domain) and \textbf{0.937} (out-of-domain) at 500 shots. Armenian shows consistent robustness for GPT-4o (\textbf{0.863} in-domain, \textbf{0.871} out-of-domain) while mistral-large achieves competitive accuracies (0.740 in-domain, 0.729 out-of-domain). Georgian demonstrates variability; mistral-large achieves the highest out-of-domain accuracy (\textbf{0.869} at 500 shots), highlighting the capability of open-weight models. Syriac remains challenging; GPT-4o reaches \textbf{0.863} (in-domain), whereas mistral-large attains the best out-of-domain accuracy (\textbf{0.728}).

In general, large language models demonstrate clear advantages over the PIE baseline, especially in low-resource scenarios. Open-weight models sometimes closely match or even surpass GPT models in certain languages or data conditions.

\section{Discussion}

Our findings highlight both the capabilities and limitations of large language models in morphosyntactic annotation tasks for under-resourced and typologically diverse languages. While traditional methods such as the RNN-based PIE baseline require substantial annotated data to achieve competitive results, LLMs—both proprietary models like GPT-4o and open-weight alternatives like mistral-large—show significant promise even in few-shot and zero-shot contexts. This ability to perform reliably with minimal supervision makes them particularly valuable for languages lacking extensive annotated resources.

\subsection{Lexical Overlap and the Generalization Gap}

To rigorously assess whether these models rely on memorization or genuine linguistic reasoning, we analyzed the lexical overlap between training and test sets (see Table \ref{tab:lexical_analysis}). A key finding is that \textit{Lemma Type Overlap} remains consistently low across all languages ($<63\%$), meaning the models must process a high volume of previously unseen vocabulary.

We observe a significant gap between \textit{Type Overlap} and \textit{Token Overlap}. In Greek, Armenian, and Georgian, token overlap is 13 to 24 percentage points higher than type overlap. This indicates that while models encounter many novel lemmas, they are primarily familiar with the most frequent lexical items. This familiarity with the "syntactic backbone" (Verbs, Nouns, and Particles) provides sufficient context to maintain robust POS-tagging accuracy ($>0.85$ F1) even on Out-Of-Vocabulary (OOV) items.

However, the Syriac (SYC) case confirms that LLMs do more than simple pattern matching. Syriac exhibits the most challenging distribution, with both type and token overlap dropping to $\approx 42\%$. In this scenario, the majority of the text consists of novel lemmata. The fact that models maintain (relatively) competitive performance despite this low "seen-ness" is a measurable indicator of their capacity for morphological generalization.

\subsection{The Impact of Polylexicality (@)}

The density of polylexical forms (marked by \texttt{@}) is a major predictor of performance degradation. Syriac presents the highest complexity: 46.5\% of its test tokens require predicting an internal segmentation boundary (e.g., \textit{PART@NOUN}). For LLMs using sub-word tokenization, generating these non-standard delimiters within non-Latin scripts is highly error-prone, directly accounting for the performance gap compared to Greek or Armenian.

Interestingly, Georgian (KAT) shows that overlap is not the only factor. Despite having the highest lexical overlap and few polylexical forms ($<2\%$), it remains more challenging than Greek. This suggests that the complexity of its agglutinative morphology—often not explicitly segmented in this schema—requires deeper linguistic pre-training that foundation models may still lack for kartvelian languages.

\subsection{Conclusion on Model Robustness}

In general, while acknowledging the limitations inherent to out-of-domain benchmarks, our results confirm that LLMs provide an efficient and cost-effective path to annotating under-resourced languages. They serve as robust "cold-start" annotators, capable of facilitating the incremental creation of gold-standard datasets for historically and linguistically diverse corpora. 

Our analysis of lexical seen-ness reveals an important distinction: while high token overlap provides a performance "floor" for frequent items, the models' ability to maintain accuracy in low-overlap environments—most notably in Syriac, where the majority of the vocabulary is novel—demonstrates a capacity for genuine morphological reasoning over simple lexical memorization.
Future work should focus on moving beyond random sampling towards strategic input selection \citep{bansal2023llmannotators}. Such strategies could further narrow the performance gap between seen and unseen tokens, significantly enhancing model generalization in the high-sparsity and domain-shift scenarios typical of under-resourced languages and non-Latin languages.

\subsection{Error Typology: Formatting vs. Linguistic Competence}

A manual review, but very limited at this stage, of the outputs reveals a fundamental distinction between structural failures and genuine morphosyntactic competence. In Syriac, the high density of polylexical forms frequently causes structural desynchronization in TSV outputs. Mismanagement of the \texttt{@} delimiter often results in lower raw accuracy scores that may not fully reflect the model's underlying linguistic competence. Similarly, in Classical Armenian, models exhibit a bias toward modern training data by hallucinating reformed orthography instead of the required classical standard.

These observations suggest that the models' true competence exceeds raw metrics, leading us to recommend the use of more robust formats, such as JSON, for future workflows to decouple data structure from philological content.

\begin{figure*}[ht!]
    \centering
    \includegraphics[width=\linewidth]{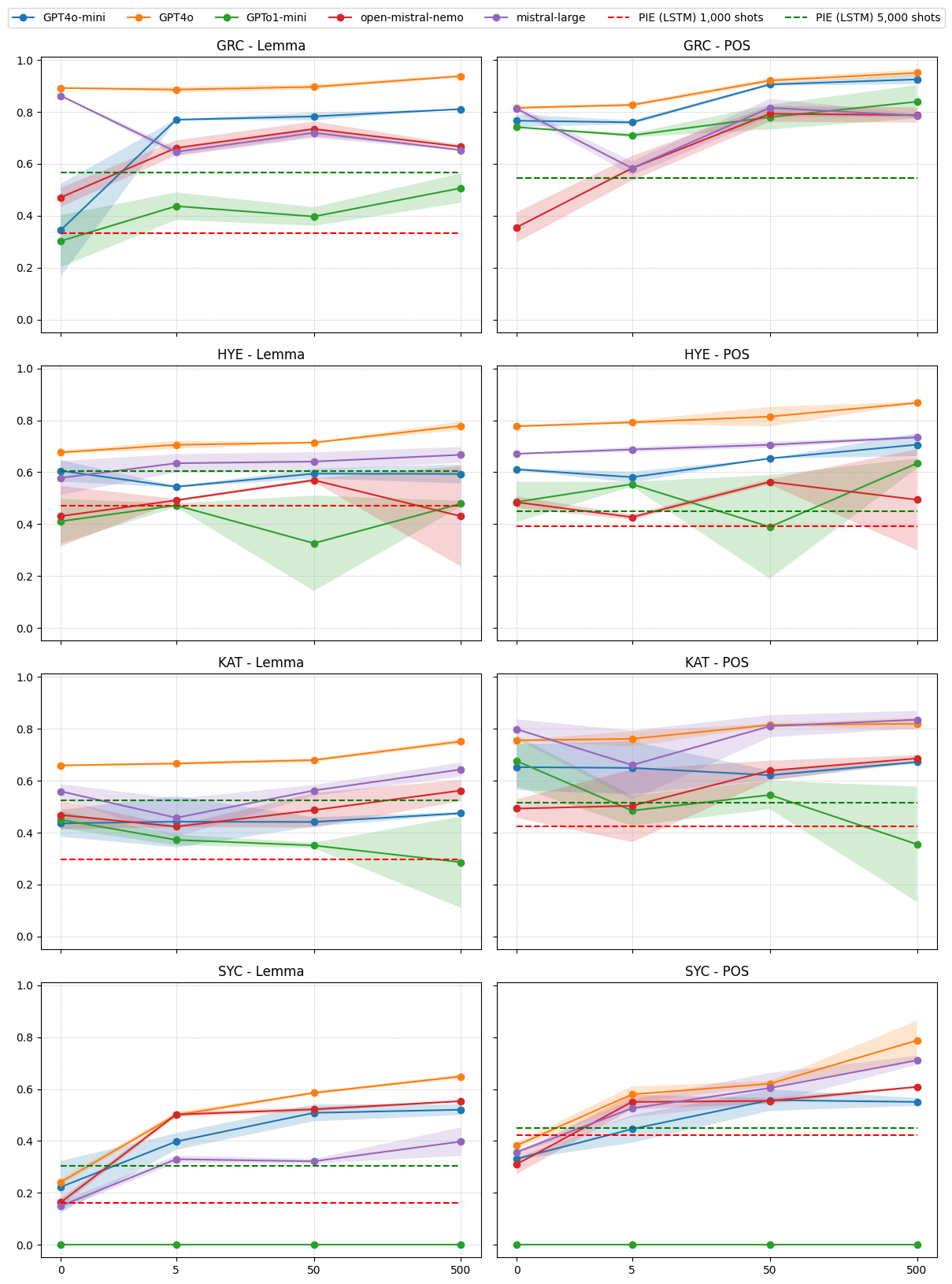}
    \caption{Figure 4. Accuracy as a function of the number of in-context examples (0, 5, 50, 500 shots) for lemmatization and POS tagging, reported separately for each language (GRC, HYE, KAT, SYC). Solid lines show the model accuracy at each shot count; shaded bands span the two evaluation conditions and thus indicate the range between in-domain and out-of-domain accuracies at the same shot count. Dashed horizontal lines mark the supervised PIE reference accuracies obtained with 1,000 and 5,000 training examples.}
    \label{fig:accuracy}
\end{figure*}

\section*{Limitations}\label{sec:limitations}
Our evaluation is constrained by the small size and genre specificity of the benchmark dataset. Although the corpora were selected to represent typological and historical diversity, each training and test set obviously remains limited in size.

Our annotation scheme, based on GRE\textit{g}ORI guidelines, introduces structural complexity through polylexical and agglutinative segmentation marked by the @ delimiter. The tagset is also highly specific to GRE\textit{g}ORI, with morphological tags within the POS list. While linguistically motivated, this representation lacks standardization, limiting cross-corpus comparability and hindering broader model development.

Model performance varies significantly across tasks and languages. LLMs showed strong results in POS tagging and lemmatization for some languages but consistently underperformed for some languages. These discrepancies suggest that few-shot generalization remains highly sensitive to script, morphology, and training distribution, with models struggling to identify entities in highly inflected, under-represented languages.

Although our experiments include a range of foundation models (GPT-4 variants and Mistral models), we do not systematically investigate the impact of prompt engineering, decoding strategies, or intermediate fine-tuning. Future work should explore how task- and language-specific prompts might further boost performance in few-shot settings.

Finally, our approach presumes the availability of token-level supervision for evaluation, which may not be feasible for other truly under-documented languages. The reliance on pre-existing annotated corpora, even if limited, underscores a key challenge in extending this approach to languages without any available training data.

\section*{Data availability}

Code, results and datasets are available at: \href{https://github.com/CVidalG/EACL2026-historical-languages}{https://github.com/CVidalG/EACL2026-historical-languages}.

\section*{Acknowledgements}
This research was supported by the French National Research Agency (ANR), grant ANR-21-CE38-0006 (DALiH project) and conducted as part of the PSL Research University's Major Research Program CultureLab, implemented by the ANR (reference ANR-10-IDEX-0001). We also thank the GRE\textit{g}ORI lab for providing access to the data. The authors used generative AI to improve the linguistic clarity and assist with LaTeX formatting of the final document.

\section*{Ethics Statement}

This research adheres to the ACL Ethics Policy. Our experiments use openly accessible linguistic datasets and publicly available LLMs, ensuring transparency and reproducibility. We acknowledge that working with historical and culturally significant texts requires sensitivity and care. 

Although our datasets consist exclusively of texts published and available in critical scholarly editions, researchers employing our methods in other contexts must ensure respect for cultural and historical heritage, particularly when annotating or disseminating results related to under-resourced or minority languages. Furthermore, we encourage future work to consider data sovereignty and community involvement when extending these techniques to languages without established digital resources.


\bibstyle{acl_natbib}
\bibliography{main}

\clearpage
\appendix
\onecolumn
\section{Appendix}

\begin{table}[htbp]
\tiny
\centering
\begin{adjustbox}{width=\textwidth}
\begin{tabular}{l|llllll||llllll}
                  & \multicolumn{6}{c}{In-domain}                                  & \multicolumn{6}{c}{Out-of-domain}                              \\\hline
Model             & 0-shot & 5-shots & 50-shots & 500-shots & 1.000-shots & 5.000-shots & 0-shot & 5-shots & 50-shots & 500-shots & 1.000-shots & 5.000-shots \\\hline
GRC               &        &        &         &          &            &            &        &        &         &          &            &            \\
GPT4o-mini        & 0.523  & 0.767  & 0.797   & 0.81     & -          & -          & 0.166  & 0.772  & 0.768   & 0.811    & -          & -          \\
GPT4o             & \textbf{0.893}  & \textbf{0.896}  & \textbf{0.906}   & \textbf{0.933}    & -          & -          & \textbf{0.891}  & \textbf{0.875 } & \textbf{0.886}   & \textbf{0.942}    & -          & -          \\
GPTo1-mini        & 0.403  & 0.49   & 0.433   & 0.563    & -          & -          & 0.202  & 0.384  & 0.361   & 0.45     & -          & -          \\
open-mistral-nemo & 0.507  & 0.69   & 0.763   & 0.673    & -          & -          & 0.434  & 0.632  & 0.705   & 0.659    & -          & -          \\
mistral-large     & 0.863  & 0.63   & 0.74    & 0.653    & -          & -          & 0.861  & 0.662  & 0.699   & 0.652    & -          & -          \\
PIE               & -      & -      & -       & 0.48     & 0.333      & 0.566      & -      & -      & -       & 0.381    & 0.301      & 0.453      \\\\
HYE               &        &        &         &          &            &            &        &        &         &          &            &            \\
GPT4o-mini        & 0.647  & 0.547  & 0.613   & 0.63     & -          & -          & 0.564  & 0.541  & 0.576   & 0.557    & -          & -          \\
GPT4o             & \textbf{0.683}  & \textbf{0.69}   & \textbf{0.716}   & \textbf{0.796}    & -          & -          & \textbf{0.67}   & \textbf{0.721}  & \textbf{0.712}   & \textbf{0.761}    & -          & -          \\
GPTo1-mini        & 0.497  & 0.48   & 0.143   & 0.49     & -          & -          & 0.325  & 0.465  & 0.51    & 0.468    & -          & -          \\
open-mistral-nemo & 0.547  & 0.497  & 0.563   & 0.237    & -          & -          & 0.315  & 0.487  & 0.576   & 0.624    & -          & -          \\
mistral-large     & 0.643  & 0.67   & 0.677   & 0.697    & -          & -          & 0.513  & 0.599  & 0.605   & 0.637    & -          & -          \\
PIE               & -      & -      & -       & 0.346    & 0.47       & 0.603      & -      & -      & -       & 0.359    & 0.468      & 0.598      \\\\
KAT               &        &        &         &          &            &            &        &        &         &          &            &            \\
GPT4o-mini        & 0.487  & 0.343  & 0.423   & 0.48     & -          & -          & 0.384  & 0.541  & 0.459   & 0.469    & -          & -          \\
GPT4o             & \textbf{0.656}  & \textbf{0.66}   & \textbf{0.686}   & \textbf{0.76}     & -          & -          & \textbf{0.661}  & \textbf{0.672}  & \textbf{0.672}   & \textbf{0.742}    & -          & -          \\
GPTo1-mini        & 0.417  & 0.393  & 0.34    & 0.11     & -          & -          & 0.482  & 0.351  & 0.361   & 0.462    & -          & -          \\
open-mistral-nemo & 0.523  & 0.42   & 0.553   & 0.603    & -          & -          & 0.413  & 0.426  & 0.42    & 0.521    & -          & -          \\
mistral-large     & 0.587  & 0.383  & 0.583   & 0.67     & -          & -          & 0.531  & 0.531  & 0.541   & 0.616    & -          & -          \\
PIE               & -      & -      & -       & 0.393    & 0.296      & 0.523      & -      & -      & -       & 0.347    & 0.265      & 0.485      \\\\
SYC               &        &        &         &          &            &            &        &        &         &          &            &            \\
GPT4o-mini        & \textbf{0.323}  & 0.366  & 0.476   & 0.500    & -          & -          & 0.123  & 0.43   & 0.541   & 0.541    & -          & -          \\
GPT4o             & 0.23   & 0.49   & \textbf{0.58}    & \textbf{0.656}    & -          & -          & \textbf{0.254}  & \textbf{0.511}  & \textbf{0.591}   & \textbf{0.641}    & -          & -          \\
GPTo1-mini        & -      & -      & -       & -        & -          & -          & -      & -      & -       & -        & -          & -          \\
open-mistral-nemo & 0.15   & \textbf{0.493}  & 0.513   & 0.556    & -          & -          & 0.178  & \textbf{0.511}  & 0.531   & 0.551    & -          & -          \\
mistral-large     & 0.166  & 0.316  & 0.313   & 0.343    & -          & -          & 0.134  & 0.344  & 0.331   & 0.453    & -          & -          \\
PIE               & -      & -      & -       & 0.186    & 0.162      & 0.305      & -      & -      & -       & 0.20     & 0.093      & 0.25          
\end{tabular}
\end{adjustbox}
\caption{Lemmatization results (best performances highlighted in bold)}
\label{tab:lemmatization-results}
\end{table}

\begin{table}[htbp]
\tiny
\centering
\begin{adjustbox}{width=\textwidth}
\begin{tabular}{l|llllll||llllll}
                  & \multicolumn{6}{c}{In-domain}                                  & \multicolumn{6}{c}{Out-of-domain}                              \\\hline
Model             & 0-shot & 5-shots & 50-shots & 500-shots & 1.000-shots & 5.000-shots & 0-shot & 5-shots & 50-shots & 500-shots & 1.000-shots & 5.000-shots \\\hline
GRC               &        &        &         &          &            &            &        &        &         &          &            &            \\
GPT4o-mini        & 0.79   & 0.767  & 0.9     & 0.943    & -          & -          & 0.742  & 0.752  & \textbf{0.911}   & 0.907    & -          & -          \\
GPT4o             & \textbf{0.82}   & \textbf{0.82}   & \textbf{0.93}    & \textbf{0.963}    & -          & -          & \textbf{0.811}  & \textbf{0.833}  & \textbf{0.911}   & \textbf{0.937}    & -          & -          \\
GPTo1-mini        & 0.74   & 0.717  & 0.827   & 0.903    & -          & -          & 0.742  & 0.702  & 0.732   & 0.775    & -          & -          \\
open-mistral-nemo & 0.413  & 0.63   & 0.823   & 0.817    & -          & -          & 0.298  & 0.536  & 0.762   & 0.758    & -          & -          \\
mistral-large     & 0.823  & 0.61   & 0.85    & 0.793    & -          & -          & 0.801  & 0.556  & 0.781   & 0.778    & -          & -          \\
PIE               & -      & -      & -       & 0.556    & 0.546      & 0.546      & -      & -      & -       & 0.46     & 0.45       & 0.476      \\\\
HYE               &        &        &         &          &            &            &        &        &         &          &            &            \\
GPT4o-mini        & 0.607  & 0.56   & 0.653   & 0.75     & -          & -          & 0.615  & 0.602  & 0.653   & 0.662    & -          & -          \\
GPT4o             & \textbf{0.773}  & \textbf{0.786}  & \textbf{0.776}   & \textbf{0.863}    & -          & -          & \textbf{0.781}  & \textbf{0.798}  & \textbf{0.852}   & \textbf{0.871}    & -          & -          \\
GPTo1-mini        & 0.563  & 0.563  & 0.19    & 0.653    & -          & -          & 0.408  & 0.545  & 0.589   & 0.615    & -          & -          \\
open-mistral-nemo & 0.507  & 0.437  & 0.553   & 0.3      & -          & -          & 0.462  & 0.417  & 0.573   & 0.688    & -          & -          \\
mistral-large     & 0.673  & 0.697  & 0.717   & 0.74     & -          & -          & 0.669  & 0.678  & 0.694   & 0.729    & -          & -          \\
PIE               & -      & -      & -       & 0.366    & 0.393      & 0.45       & -      & -      & -       & 0.35     & 0.379      & 0.461      \\\\
KAT               &        &        &         &          &            &            &        &        &         &          &            &            \\
GPT4o-mini        & 0.74   & 0.547  & 0.603   & 0.677    & -          & -          & 0.564  & 0.751  & 0.639   & 0.666    & -          & -          \\
GPT4o             & \textbf{0.756}  & \textbf{0.79}   & \textbf{0.82}    & 0.796    & -          & -          & 0.754  & 0.732  & 0.809   & 0.842    & -          & -          \\
GPTo1-mini        & 0.577  & 0.543  & 0.49    & 0.133    & -          & -          & 0.774  & 0.426  & 0.6     & 0.577    & -          & -          \\
open-mistral-nemo & 0.457  & 0.363  & 0.677   & 0.673    & -          & -          & 0.528  & 0.643  & 0.6     & 0.698    & -          & -          \\
mistral-large     & 0.76   & 0.527  & 0.767   & \textbf{0.8}      & -          & -          & \textbf{0.836}  & \textbf{0.793}  & \textbf{0.852}   & \textbf{0.869}    & -          & -          \\
PIE               & -      & -      & -       & 0.39     & 0.423      & 0.516      & -      & -      & -       & 0.403    & 0.406      & 0.561      \\\\
SYC               &        &        &         &          &            &            &        &        &         &          &            &            \\
GPT4o-mini        & 0.327  & 0.393  & 0.516   & 0.566    & -          & -          & 0.336  & 0.499  & 0.599   & 0.535    & -          & -          \\
GPT4o             & 0.37   & \textbf{0.61}   & \textbf{0.61}    & \textbf{0.863}    & -          & -          & \textbf{0.396}  & 0.549  & \textbf{0.631}   & 0.711    & -          & -          \\
GPTo1-mini        & \textbf{0.513}  & 0.513  & 0.53    & 0.513    & -          & -          & 0.465  & 0.179  & 0.163   & 0.429    & -          & -          \\
open-mistral-nemo & 0.35   & 0.523  & 0.566   & 0.606    & -          & -          & 0.272  & \textbf{0.578}  & 0.543   & 0.611    & -          & -          \\
mistral-large     & 0.353  & 0.56   & 0.546   & 0.693    & -          & -          & 0.362  & 0.492  & 0.662   & \textbf{0.728}    & -          & -          \\
PIE               & -      & -      & -       & 0.196    & 0.423      & 0.45       & -      & -      & -       & 0.235    & 0.431      & 0.518     
\end{tabular}
\end{adjustbox}
\caption{POS-tagging results (best performances highlighted in bold)}
\label{tab:pos-results}
\end{table}

\begin{table*}[htbp]
\small
\centering
\begin{tabular}{l|l|cccc}
Tag & Explanation & GRC & HYE & KAT & SYC \\\hline

A & Adjective & \checkmark & \checkmark & \checkmark & \checkmark \\
ADV & Adverb & & & & \checkmark \\
AMORPH & Element of morphological analysis & \checkmark & & & \\
CARD & Cardinal Number & & & & \checkmark \\
DET & Article & \checkmark & & & \\
ETYM & Etymon & \checkmark & & & \\
I+Adv & Adverb & \checkmark & \checkmark & \checkmark & \\
I+AdvPr & Prepositional Adverb & \checkmark & \checkmark & & \\
I+Conj & Conjunction & \checkmark & \checkmark & \checkmark & \\
I+Intj & Interjection & \checkmark & \checkmark & \checkmark & \\
I+Neg & Negation & \checkmark & \checkmark & & \\
I+Part & Particle & \checkmark & \checkmark & \checkmark & \\
I+Prep & Preposition & \checkmark & \checkmark & \checkmark & \\
LF & Unanalyzable form chosen as lemma (lemma-form) & \checkmark & \checkmark & \checkmark & \\
N+Ant & Anthroponymic Name & \checkmark & \checkmark & \checkmark & \\
N+Com & Common Noun & \checkmark & \checkmark & \checkmark & \\
N+Epi & Epiclesis (Nickname) & \checkmark & \checkmark & & \\
N+Lettre & Name of a letter & \checkmark & & & \\
N+Let & Name of a letter & & \checkmark & & \\
N+Pat & Patronymic Name & \checkmark & \checkmark & \checkmark & \\
N+Prop & Proper Noun & \checkmark & \checkmark & \checkmark & \\
N+Top & Toponym (Place Name) & \checkmark & \checkmark & \checkmark & \\
NAME & Proper Name & & & & \checkmark \\
NAME\_ant & Anthroponymic Name & & & & \checkmark \\
NAME\_top & Toponymic Name & & & & \checkmark \\
NUM+Car & Cardinal Number (word) & \checkmark & \checkmark & \checkmark & \\
NUM+Ord & Ordinal Number (word) & \checkmark & \checkmark & \checkmark & \\
NUMA+Car & Cardinal Number (alphanumeric system) & \checkmark & \checkmark & \checkmark & \\
NUMA+Ord & Ordinal Number (alphanumeric system) & \checkmark & \checkmark & \checkmark & \\
ORD & Ordinal Number & & & & \checkmark \\
PART & Particle & & & & \checkmark \\
PRO+Dem & Demonstrative Pronoun & \checkmark & \checkmark & \checkmark & \\
PRO+Ind & Indefinite Pronoun & \checkmark & \checkmark & \checkmark & \\
PRO+Int & Interrogative Pronoun & \checkmark & \checkmark & \checkmark & \\
PRO+Per1d & Personal Pronoun 1st Person Dual & \checkmark & & & \\
PRO+Per1p & Personal Pronoun 1st Person Plural & \checkmark & \checkmark & \checkmark & \\
PRO+Per1s & Personal Pronoun 1st Person Singular & \checkmark & \checkmark & \checkmark & \\
PRO+Per2p & Personal Pronoun 2nd Person Plural & \checkmark & \checkmark & \checkmark & \\
PRO+Per2s & Personal Pronoun 2nd Person Singular & \checkmark & \checkmark & \checkmark & \\
PRO+Per3p & Personal Pronoun 3rd Person Plural & \checkmark & & & \\
PRO+Per3s & Personal Pronoun 3rd Person Singular & \checkmark & & & \\
PRO+Pos1p & Possessive Pronoun 1st Person Plural & \checkmark & \checkmark & \checkmark & \\
PRO+Pos1s & Possessive Pronoun 1st Person Singular & \checkmark & \checkmark & \checkmark & \\
PRO+Pos2p & Possessive Pronoun 2nd Person Plural & \checkmark & \checkmark & \checkmark & \\
PRO+Pos2s & Possessive Pronoun 2nd Person Singular & \checkmark & \checkmark & \checkmark & \\
PRO+Pos3p & Possessive Pronoun 3rd Person Plural & \checkmark & & \checkmark & \\
PRO+Pos3s & Possessive Pronoun 3rd Person Singular & \checkmark & & \checkmark & \\
PRO+Rec & Reciprocal Pronoun & \checkmark & \checkmark & \checkmark & \\
PRO+Ref & Reflexive Pronoun & & \checkmark & & \\
PRO+Ref1s & Reflexive Pronoun 1st Person Singular & \checkmark & & & \\
PRO+Ref2s & Reflexive Pronoun 2nd Person Singular & \checkmark & & & \\
PRO+Ref3s & Reflexive Pronoun 3rd Person Singular & \checkmark & & & \\
PRO+Rel & Relative Pronoun & \checkmark & \checkmark & \checkmark & \\
PRO\_dem & Demonstrative Pronoun & & & & \checkmark \\
PRO\_ind & Indefinite Pronoun & & & & \checkmark \\
PRO\_int & Interrogative Pronoun & & & & \checkmark \\
PRO\_pers & Personal Pronoun & & & & \checkmark \\
V & Verb & \checkmark & \checkmark & & \\
V+Mas & Mazdar Verb & & & \checkmark & \\
V+Part & Participial Verb & & & \checkmark & \\
V1--V33 & Various Syriac verb forms (Pe'al, Pa''el, etc.) & & & & \checkmark \\
\end{tabular}
\caption{List of POS-tags in GRE\textit{g}ORI datasets}
\label{tab:pos-list}
\end{table*}



\end{document}